\documentclass{article} % For LaTeX2e
\usepackage{iclr2015,times}
\usepackage{hyperref}
\usepackage{url}
\usepackage{amsthm,amssymb,amsfonts,amsmath}
\usepackage{graphicx}
\usepackage{caption}
\usepackage{subcaption}

\title{Language Recognition using Random Indexing}

\author{
Aditya Joshi \\
Department of Mathematics\\
University of California, Berkeley\\
Berkeley, CA 94720, USA \\
\texttt{adityajoshi@berkeley.edu} \\
\And
Johan T. Halseth \\
Department of Computer Science \\
University of California, Berkeley\\
Berkeley, CA 94720, USA \\
\texttt{halseth@berkeley.edu} \\
\AND
Pentti Kanerva\\
Redwood Center for Theoretical Neuroscience\\
University of California, Berkeley\\
Berkeley, CA 94720, USA \\
\texttt{pkanerva@berkeley.edu}
}

\iclrconference % Uncomment if submitted as conference paper instead of workshop

\begin{document}

\maketitle

\begin{abstract}
Random Indexing is a simple implementation of Random Projections with
a wide range of applications. It can solve a variety of problems with
good accuracy without introducing much complexity. Here we demonstrate
its use for identifying the language of text samples, based on a novel
method of encoding letter $n$-grams into high-dimensional Language
Vectors.  Further, we show that the method is easily implemented and
requires little computational power and space.  As proof of the
method's statistical validity, we show its success in a
language-recognition task. On a difficult data set of 21,000 short
sentences from 21 different languages, we achieve 97.8\% accuracy,
comparable to state-of-the-art methods.
\end{abstract}

\section{Introduction}

As humans who communicate through language, we have the fascinating
ability to recognize unknown languages in spoken or written form,
using simple cues to distinguish one language from another. Some
unfamiliar languages, of course, might sound very similar, especially
if they come from the same language family, but we are often able to
identify the language in question with very high accuracy. This is
because embedded within each language are certain features that
clearly distinguish one from another, whether it be accent, rhythm, or
pitch patterns. The same can be said for written languages, as they
all have features that are unique. Recognizing the language of a given
text is the first step in all sorts of language processing, such as
text analysis, categorization, translation and much more.

%\\ \\

As popularized by \cite{shannon}, most language models use
distributional statistics to explain structural similarities in
various specified languages. The traditional method of identifying
languages consists of counting individual letters, letter bigrams,
trigrams, tetragrams, etc., and comparing the frequency profiles of
different text samples. As a general principle, the more accurate you
want your detection method to be, the more data you have to store
about the various languages. For example, Google's recently
open-sourced program called Chromium Compact Language Detector uses
large language profiles built from enormous corpora of data. As a
result, the accuracy of their detection, as seen through large-scale
testing and in practice, is near perfect (\cite{mccandless}).

%\\ \\

High-dimensional vector models are popular in natural-language
processing and are used to capture word meaning from word-use
statistics.  The vectors are called {\it semantic vectors} or {\it
  context vectors}.  Ideally, words with a similar meaning are
represented by semantic vectors that are close to each other in the
vector space, while dissimilar meanings are represented by semantic
vectors far from each other. Latent Semantic Analysis is a well-known
model that is explained in detail in \cite{landauer}.  It produces
300-dimensional (more or less) semantic vectors from a singular value
decomposition (SVD) of a matrix of word frequencies in a large
collection of documents.

An alternative to SVD, based on Random Projections, was proposed by
\cite{papadimitriou} and \cite{kaski}.  Random Indexing
(\cite{kanervakristoferson,sahlgren}) is a simple and effective
implementation of the idea.  It has been used in ways similar to
Mikolov et al.'s Continuous Bag-of-Words Model (KBOW; \cite{mikolov})
and has features similar to Locality-Sensitive Hashing (LSH) but
differs from them in its use of high dimensionality and randomness.
With the dimensionality in the thousands (e.g., $D$ =
10,000)---referred to as ``hyperdimensional''---it is possible to
calculate useful representations in a {\it single pass} over the
dataset with very little computing.

%\\ \\

In this paper, we will present a way of doing language detection using
Random Indexing, which is fast, highly scalable, and space efficient.
We will also present some results regarding the accuracy of the
method, even though this will not be the main goal of this paper and
should be investigated further.

\section{Random Indexing}

Random Indexing stores information by projecting data onto vectors in
a hyperdimensional space. There exist a huge number of different,
nearly orthogonal vectors in such a space \cite[p.~19]{kanervasparse}.
This lets us combine two such vectors into a new vector using
well-defined vector-space operations, while keeping the information of
the two with high probability. In our implementation of Random
Indexing, we use a variant of the MAP (Multiply, Add, Permute) coding
described in \cite{levy09} to define the hyperdimensional vector
space. Vectors are initially taken from a $D$-dimensional space (with
$D$ = 10,000) and have an equal number of randomly placed 1s and \,
$-1$s.  Such vectors are used to represent the basic elements of the
system, which in our case are the 26 letters of the alphabet and the
(ASCII) Space.  These vectors for letters are sometimes referred to as
their {\it Random Labels}.

%\\ \\

The binary operations on such vectors are defined as follows.
Elementwise {\it addition} of two vectors $A$ and $B$, is denoted by
$A + B$. Similar, elementwise {\it multiplication} is denoted by $A *
B$.  A vector $A$ will be its own multiplicative inverse, $A * A$ =
{\bf 1}, where {\bf 1} is the $D$-dimensional identity vector
consisting of only $1$s. Cosine angles are used to measure the
similarity of two vectors.  It is defined as $\cos(A,B) = |A' * B'|$,
where $A'$ and $B'$ are the normalized vectors of $A$ and $B$,
respectively, and $|C|$ denotes the sum of the elements in $C$.

%\\ \\

Information from a pair of vectors $A$ and $B$ is stored and utilized
in a single vector by exploiting the summation operation. That is, the
sum of two separate vectors naturally preserves unique information
from each vector because of the mathematical properties of the
hyperdimensional space.  To see this, note that $\cos(A,A) = 1$, while
for all $B \neq A$, $\cos(A,B) < 1$. The cosine of two random,
unrelated vectors tend to be close to $0$. Because of this, the vector
$B$ can easily be found in the vector $A+B$: $\cos(B, A + B)$ differs
significantly from 0.

%\\ \\

For storing sequences of vectors, we use a random (but fixed
throughout all our computations) {\it permutation} operation $\rho$ of
the vector coordinates. Hence, the sequence A-B-C, is stored as the
vector $(\rho((\rho A) * B )) * C = \rho \rho A * \rho B * C$. This
efficiently distinguishes the sequence A-B-C from, say, A-C-B. This
can be seen from looking at their cosine (here $c$ is the
normalization factor):
\begin{align*}
	V_1 &= \rho \rho A * \rho B * C \\
    V_2 &= \rho \rho A * \rho C * B \\
    \implies \cos(V_1, V_2) &= c \cdot |(\rho \rho A * \rho B * C) * (\rho \rho A * \rho C * B)| \\
    &= c \cdot |\rho \rho A * \rho \rho A * \rho B * \rho C * C  * B)| \\
    &= c \cdot |\rho \rho (A * A) * \rho (B * C) * (B * C))| \\
    &\approx c \cdot 0
\end{align*}
since a random permutation $\rho V_1$ of a random vector $V_1$ is
uncorrelated to $V_2$.

\subsection{Making and Comparing of Text Vectors}

We use the properties of hyperdimensional vectors to extract certain
properties of text into a single vector. \cite{kanerva10kwords} shows
how Random Indexing can be used for representing the contexts in which
a word appears in a text, into that word's context vector. We show
here how to use a similar strategy for recognizing a text's language
by creating and comparing \emph{Text Vectors}: the Text Vector of an
unknown text sample is compared for similarity to precomputed Text
Vectors of known language samples---the latter are referred to as
\emph{Language Vectors}.

%\\ \\

Simple language recognition can be done by comparing letter
frequencies of a given text to known letter frequencies of languages.
Given enough text, a text's letter distribution will approach the
letter distribution of the language in which the text was written.
The phenomenon is called an ``ergodic'' process in \cite{shannon}, as
borrowed from similar ideas in physics and thermodynamics. This can be
generalized to using \emph{letter blocks} of different sizes. By a
block of size $n$, we mean $n$ consecutive letters in the text so that
a text of length $m$ would have $m - n + 3$ blocks.  When the letters
are taken in the order in which they appear in the text, they are
referred to as a sequences (of length $n$) or as $n$-grams.

%\\ \\

As an example, the text ``a brook'' gives rise to the tetragrams
``--a--b'', ``a--br'', ``--bro'', ``broo'', ``rook'', and ``ook--''
(here ``--'' stands for Space).  The frequencies of such letter blocks
can be found for a text and compared to known frequencies for
different languages. For texts in languages using the Latin alphabet
of $26$ letters (plus Space), like English, this would lead to keeping
track of $27^4$ = 531,441 different tetragram frequencies. For
arbitrary alphabets of $l$ letters, there would be $(l+1)^n$ $n$-grams
to keep track of.  These numbers grow quickly as the block size $n$
increases.

%\\ \\

The Random Indexing approach for doing language recognition is
similar. A text's Text Vector is first calculated by running over all
the blocks of size $n$ within the text and creating an $n$-Gram Vector
for each. An $n$-Gram Vector is created for the sequence of letters as
described earlier. As an example, if we encounter the block ``grab'',
its vector is calculated by performing $\rho \rho \rho G + \rho \rho R
+ \rho A + B$, where $G$, $R$, $A$ and $B$ are the Random Labels for
g, r, a, and b---they are random $D$-dimensional vector with half $1$s
and half $-1$s and they remain constant.

%\\ \\

A text's Text Vector is now obtained from summing the $n$-Gram Vectors
for all the blocks in the text. This is still an $D$-dimensional
vector and can be stored efficiently. Language Vectors are made in
exactly the same way, by making Text Vectors from samples of a known
language and adding them into a single vector. Determining the
language of an unknown text is done by comparing its Text Vector to
all the Language Vectors.  More precisely, the cosine angle measure
$d_{\cos}$ between a language vector $X$ and an unknown text vector
$V$ is defined as follows:
\begin{align*}
d_{\cos}(X,V) = \frac{X \cdot V}{|X||V|} = \frac{\sum_{i=1}^D x_i
  v_i}{\sqrt{\sum_{j=1}^D x_j^2 \sum_{k=1}^D v_k^2}}
\end{align*}
If the cosine angle is high (close to 1), the block frequencies of the
text are similar to the block frequencies of that language and thus,
the text is likely to be written in the same language.  Hence, the
language that yields the highest cosine is chosen as the system's
prediction/guess.

%\begin{figure}[!htb]
%\centering
%\includegraphics[width=0.7\textwidth]{Nk_contours-ridiculous_D.png}
%\caption{Choosing $D$ from 1000 to 10000 and $k$ from $1$ to $10^4$,
%  we find that the variance of the similarity data is greater for
%  larger $D$ and larger $k$. Language vectors were generated with
%  ordered block sizes 2, 3, and 4 and were all added together to form
%  a multi-block language vector representation. Languages tested here
%  were English, German, Norwegian, Finnish, Dutch, French, Afrikaans,
%  Danish, and Spanish.}
%\end{figure}

\subsection{Complexity}

The outlined algorithm for Text Vector generation can be implemented
efficiently. For generating a vector for an $n$-gram, $n-1$ vector
additions and permutations are performed. This takes time $O(n \cdot
D)$. Looping over a text of $m$ letters, $O(m)$ $n$-Gram Vectors must
be created and added together. This clearly implies an $O(n \cdot D
\cdot m)$ implementation. This can be improved to $O(D \cdot m)$ by
noting that most of the information needed for creating the $n$-Gram
Vector for the next block is already contained in the previous
$n$-Gram Vector, and can be retrieved by removing the contribution
from the letter that is now no longer in the block.

%\\ \\

Say we have the $n$-Gram Vector $A = \rho^{(n-1)} V_1 * \rho^{(n-2)}
V_2 * \ldots * \rho V_{n-1} * V_n $ for block number $i$, and now want
to find the $n$-Gram Vector $B$ for block $i+1$. We remove from $A$
the vector $\rho^{(n)} V_1$ by multiplying with its inverse (which is
the vector itself), which we can do in $O(D)$ time since
$\rho^{(n-1)}$ is just another (pre-calculated) permutation. Then we
permute the result once using $\rho$ and multiply that with the Letter
Vector $V_{n+1}$ for the new letter in the block. This gives us the
new $n$-Gram Vector
\begin{align*}
	B &= \rho(\rho^{(n-1)} V_1 * A) * V_{n+1} \\
    &= \rho(\rho^{(n-2)} V_2 * \ldots *  \rho V_{n-1} * V_n) * V_{n+1} \\
    &= \rho^{(n-1)} V_2 * \ldots * \rho^{(2)}V_{n-1} * \rho V_n * V_{n+1}
\end{align*}
and so we can create $n$-Gram Vectors for arbitrary size blocks
without adding complexity.  In practice it means that the text is
processed in a single pass at about a 100,000 letters a second on a
laptop computer.

\begin{figure}%[!htb]
%\centering
%\includegraphics[width=\textwidth]{lang_map2.png}
%\caption{Language vectors roughly cluster based on their known
%relations to other languages. To create this plot, we fixed $N=10000$
%and $k=10$ and generated multi-block language vectors of ordered
%block size 1 and 2 and added history vectors too. Hyperdimensional
%language vectors were projected onto a 2 dimensional space using
%t-sne \cite{tsne}.}
        \centering
%        \begin{subfigure}[b]{\textwidth}
%        		\includegraphics[width=\textwidth]{lang_map1_bigger.png}
%				\caption{Sparse vectors.}
%        \end{subfigure}
        \begin{subfigure}[b]{\textwidth}
        		\includegraphics[width=\textwidth]{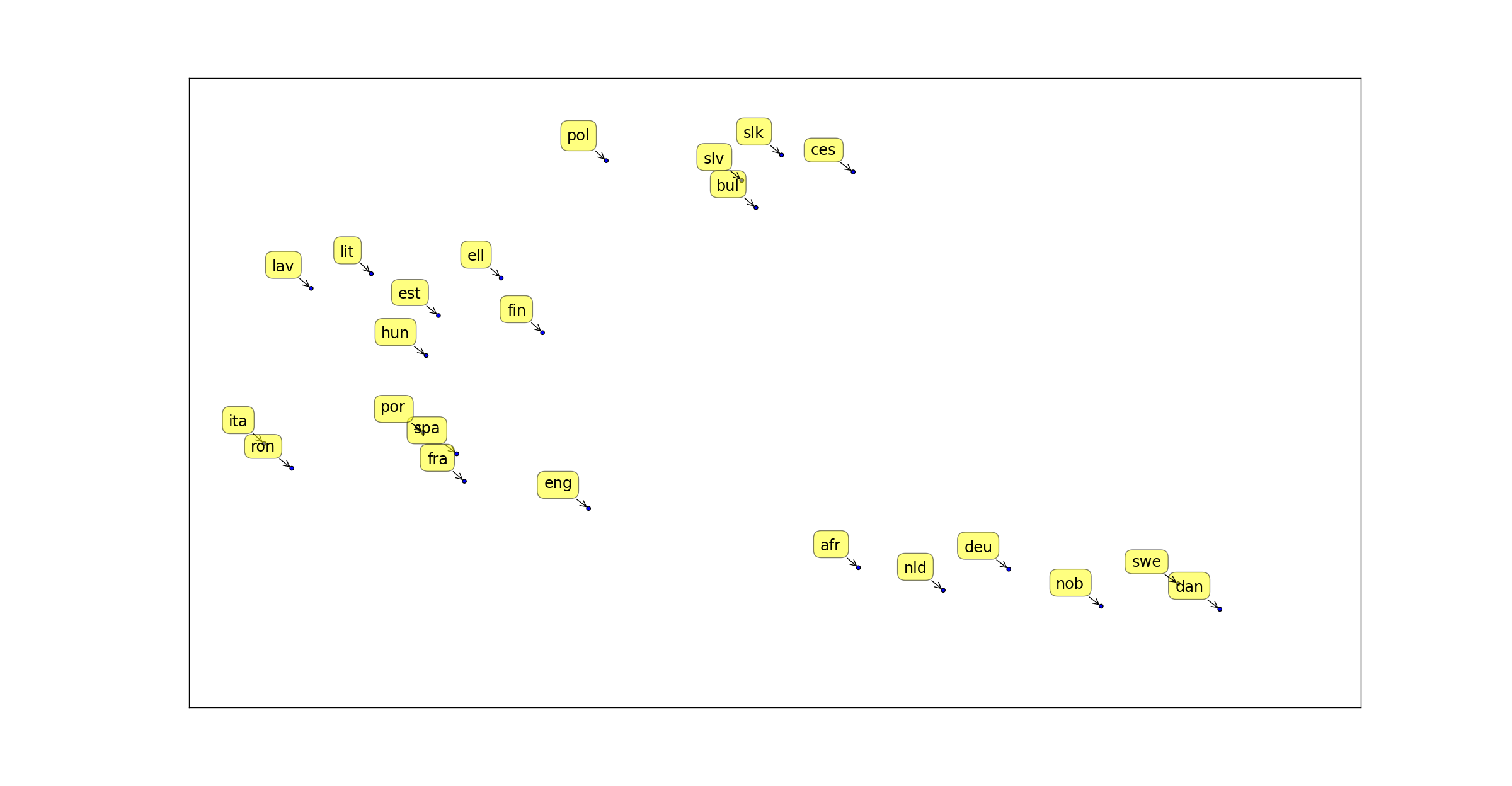}
        \end{subfigure}
        \caption*{{\bf Figure 1}: 10,000-dimensional Language Vectors
          for 23 languages roughly cluster based on the known
          relations between the languages. The Language Vectors were
          based on letter trigrams and were projected onto a plane
          using t-sne (\cite{tsne}).}
          \label{fig:lang_maps}
\end{figure}

\section{Experimental Results}

The algorithm outlined above was implemented by (\cite{github}), and
used to create Language Vectors for 23 languages.  Texts for the
Language Vectors were taken from Project Gutenberg
(\cite{projectgutenberg}) where it is available in a number of
languages, and from the Wortschatz Corpora (\cite{wortschatz}) where
large numbers of sentences in selected languages can be easily
downloaded. Each Language Vector was based on about 100,000 bytes of
text.  Computing of the Language Vectors corresponds to training the
system and took about 1 second per language on a laptop computer.

%\\ \\

Intuitively, Language Vectors within a language family should be
closer to each other than vectors for unrelated languages. Indeed, the
hyperdimensional Language Vectors roughly cluster in this manner, as
seen in Figure 1.

%\\ \\

To get an idea of how well the actual detection algorithm works, we
tested the Language Vectors' ability to identify text samples from the
Europarl Parallel Corpus, described in \cite{shuyo}.  This corpus
includes 21 languages with 1,000 samples of each, and each sample is a
single sentence.

%\\ \\

Table 1 shows the result for $n$-gram sizes from 1 to 5 ($n = 1$ is
the equivalent of comparing letter histograms).  With tetragrams we
were able to guess the correct language with 97.8\% accuracy.  Even
when incorrect, the system usually chose a language from the same
family, as seen from Table 2.

% \newpage
% \clearpage

\begin{table}
        \centering
%%        \begin{subfigure}[b]{0.45\textwidth}
                \begin{tabular}{c|l}
			$n$ & Detection success \\
			\hline
			  1 & 74.9 \\
			  2 & 94,0 \\
			  3 & 97.3 \\
			  4 & 97.8 \\
			  5 & 97.3 \\
		\end{tabular}
%%        \end{subfigure}
        \caption*{{\bf Table 1}: Percentage of sentences correctly identified
           as a function of $n$-gram size.}\label{fig:table1}
\end{table}

\begin{table}[!htb]
\centering
%% \hspace*{-1.9cm}\includegraphics[width=1.4\textwidth]{confusion_matrix2.png}

\hspace*{-2.1cm}
 \includegraphics [width=1.4\textwidth] {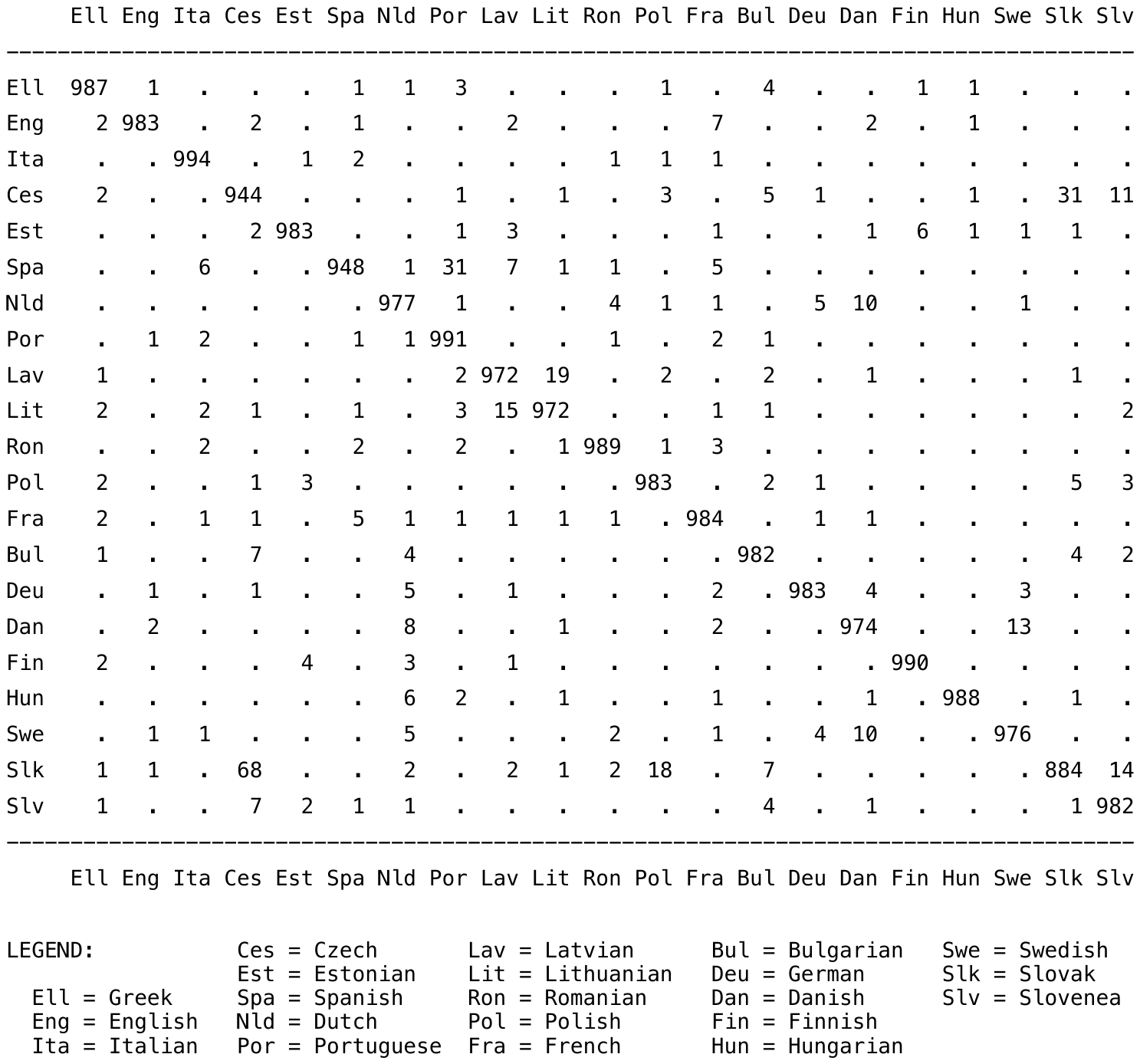}

\vspace*{-12.2cm}

\caption*{{\bf Table 2}: The confusion matrix of language detection
  using 10,000-dimensional Language Vectors based on letter trigrams.
  Each row corresponds to the correct label and each column is the
  predicted label for the Europarl corpus detection test. The entry
  $(i,j)$ is the number of sentences (out of a 1,000) that language
  $j$ was guessed for language $i$. A high value diagonal shows the
  very high accuracy.}

\end{table}

It is worth noting that 10,000-dimensional Language Vectors are
keeping track of 531,441 possible tetragrams and easily accommodate
pentagrams and beyond if needed.  The method should be explored
further, as explained in the Future Work section.

%\\ \\

%\begin{table}[t]
%\caption{Sample table title}
%\label{sample-table}
%\begin{center}
%\begin{tabular}{ll}
%\multicolumn{1}{c}{\bf PART}  &\multicolumn{1}{c}{\bf DESCRIPTION}
%\\ \hline \\
%Dendrite         &Input terminal \\
%Axon             &Output terminal \\
%Soma             &Cell body (contains cell nucleus) \\
%\end{tabular}
%\end{center}
%\end{table}

\section{Future Work}

Many adjustments can be made to improve the efficacy of Random
Indexing on language detection.  The results of this paper are based
solely on letter trigrams and tetragrams.  However, it is a simple
matter to add into the Text Vectors single-letter frequencies and
bigrams, for example.  Also, the vector dimensionality can be reduced
to several thousands without markedly affecting the results.

The arithmetic (algebra) of the operations with which Text Vectors are
made---i.e., permutation, multiplication, and addition, and how they
work together---make it possible to analyze the Language Vectors and
find out, for example, what letters are most likely to follow ``the''.
(In English it would be the Space, but what is the next most likely?)
Notice that we don't need to contemplate such questions in advance and
then design the data-gathering algorithm with that in mind.  The
information is in the vectors in a form that allows it to be retrieved
with the arithmetic.

%\\ \\

Because of the generality of Random Indexing on texts, any time series
with a well-defined ``alphabet'' can be encoded using this scheme. In
this way, we propose that our method can be used to do language
detection in speech data, addressing our original problem.

\section{Conclusion}

We have described the use of Random Indexing to language
identification.  Random Indexing has been used in the study of
semantic vectors since 2000 (\cite{kanervakristoferson,sahlgren}), and
for encoding problems in graph theory (\cite{levy09}), but only now
for identifying source materials.  It is based on simple operations on
high-dimensional random vectors: on Random Labels with 0-mean
components that allow weak signals to rise above noise as the data
accumulate.  The algorithm works in a single pass, in linear time,
with limited memory, and thus is inherently scalable, and it produces
vectors that are amenable to further analysis.  The experiments
reported in this paper were an easy task for a laptop computer.

\subsubsection*{Acknowledgments}

We would like to thank Bruno Olshausen, Mayur Mudigonda, and many
others at the Redwood Center for Theoretical Neuroscience for
insightful discussions and feedback.

% \newpage
% \clearpage

\bibliography{iclr2015}
\bibliographystyle{iclr2015}

\end{document}